\documentclass[runningheads, envcountsame, a4paper]{llncs}
\usepackage{booktabs}
\usepackage{multirow}
\usepackage{tabularx}
\usepackage{subcaption}
\usepackage{microtype}
\usepackage[pdftex]{graphicx}
\usepackage{epstopdf}
\usepackage[misc]{ifsym}

\usepackage{amsthm}

\usepackage{algorithm}
\usepackage{algorithmicx}
\usepackage[noend]{algpseudocode}

\usepackage{tikz}

\usepackage{array}
\usepackage{amsfonts}
\usepackage{amssymb}
\usepackage{amsmath}

\DeclareMathOperator*{\argmin}{arg\,min}

\newcommand*\Input[1]{\Statex \textbf{Input:} #1}
\newcommand*\Output[1]{\Statex \textbf{Output:} #1}
\algrenewcommand\alglinenumber[1]{#1}

\newcolumntype{M}[1]{>{\centering\arraybackslash}m{#1}}

\begin{document}
\definecolor{db}{RGB}{0, 0, 83}
\newcommand*\circled[1]{\tikz[baseline=(char.base)]{
            \node[shape=circle,fill=db,draw,inner sep=0.5pt] (char) {\textsf{{\color{white}#1}}};}}

\newtheoremstyle{mystyle}
  {\abovedisplayskip}     
  {\belowdisplayskip}     
  {}                      
  {}                      
  {\bfseries}             
  {.}                     
  {.5em}                  
  {}                      

\theoremstyle{mystyle}
\newtheorem{define}{Definition}

\newlength{\mheight}
\settoheight{\mheight}{M}

\newcommand{\iconthiswork}{
  \includegraphics[height=\mheight]{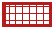}
}

\newcommand{\iconnosep}{
  \includegraphics[height=\mheight]{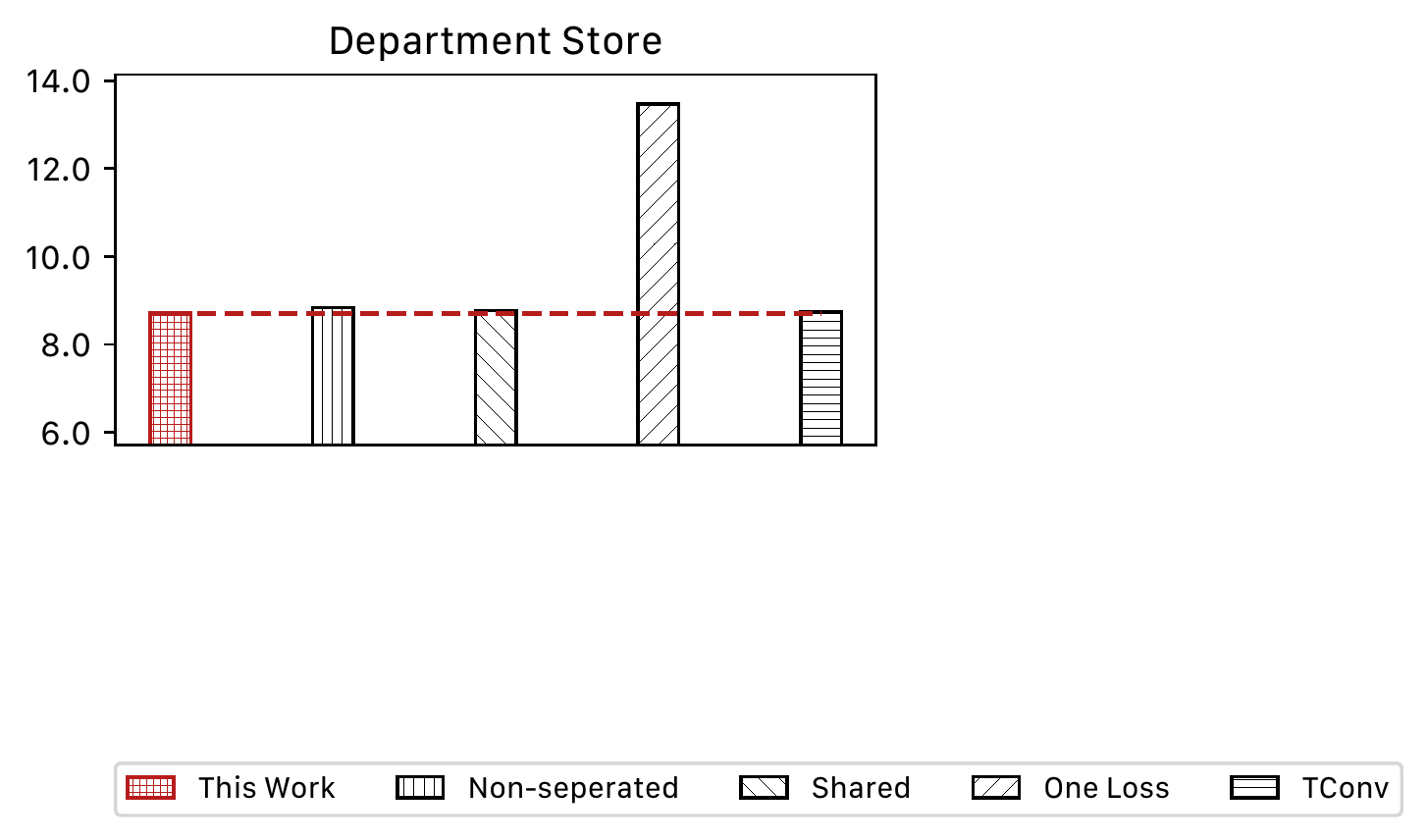}
}

\newcommand{\icononeloss}{
  \includegraphics[height=\mheight]{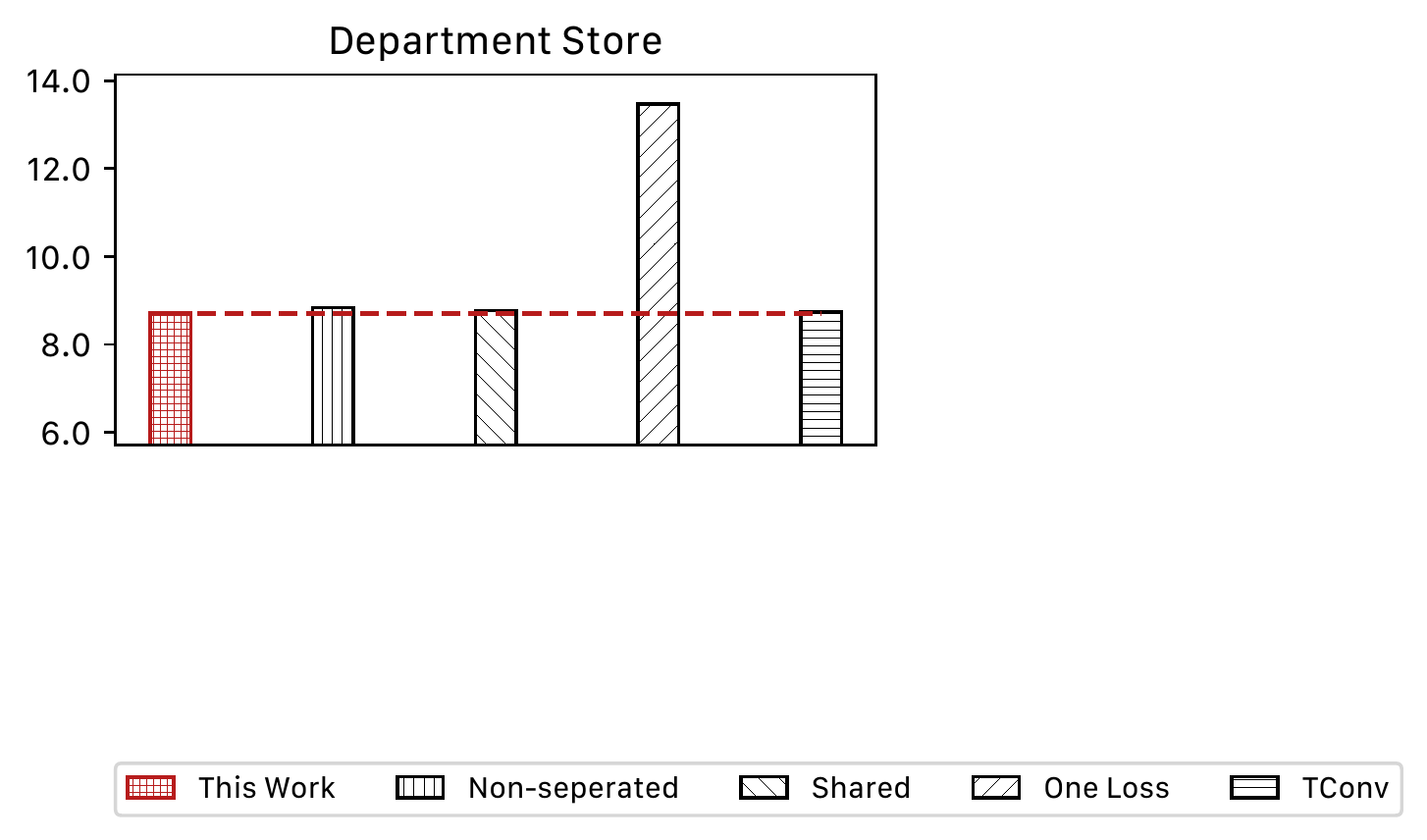}
}

\newcommand{\iconshared}{
  \includegraphics[height=\mheight]{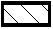}
}

\newcommand{\icontcov}{
  \includegraphics[height=\mheight]{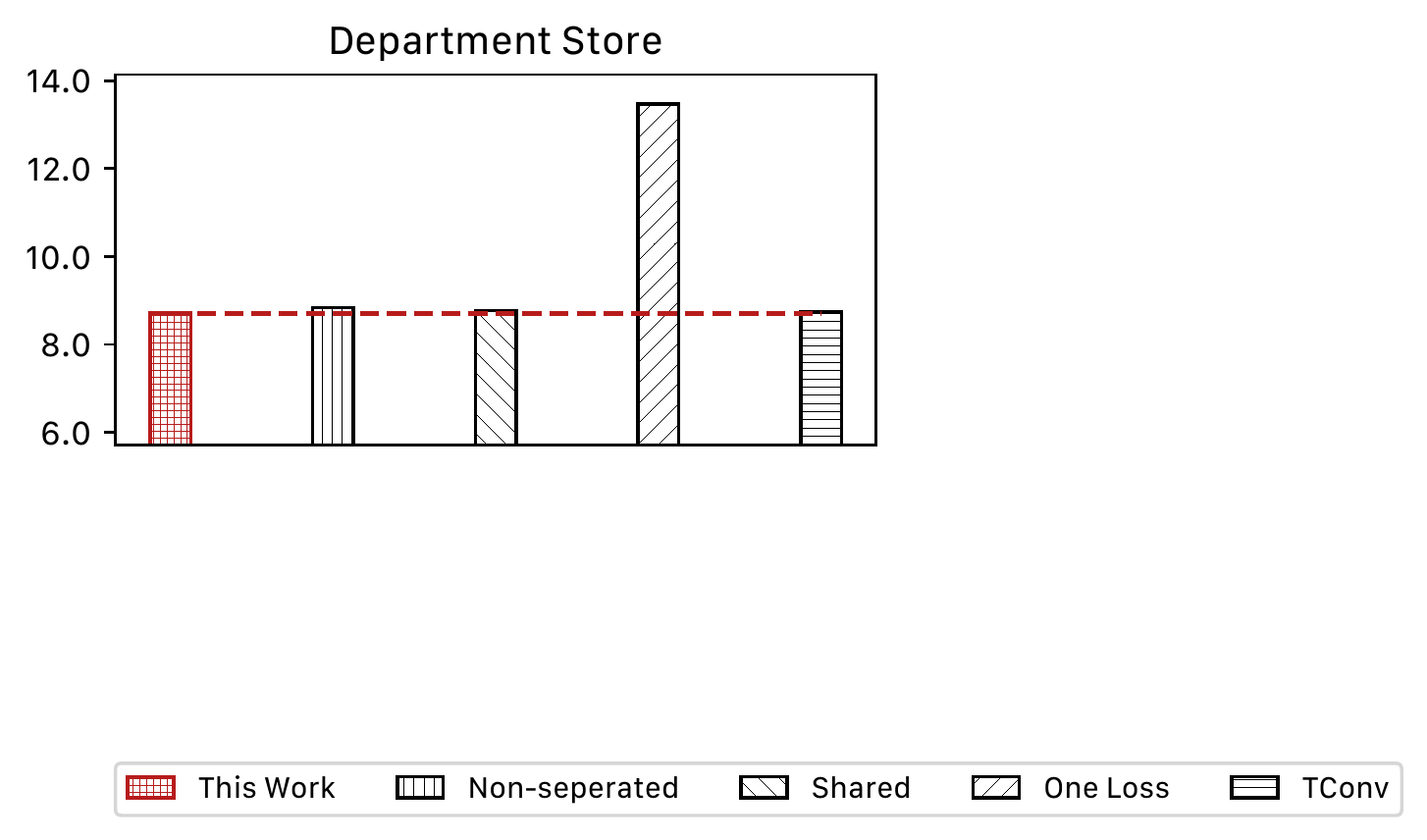}
}
\toctitle{Multi-future Merchant Transaction Prediction}
\tocauthor{Chin-Chia~Michael~Yeh, Zhongfang~Zhuang, Wei~Zhang, Liang~Wang}

\title{Multi-future Merchant Transaction Prediction}

\authorrunning{C.-C. M. Yeh, Z. Zhuang, W. Zhang, and L. Wang}
\author{Chin-Chia Michael Yeh(\Letter), Zhongfang Zhuang(\Letter), Wei Zhang \and Liang Wang
\institute{Visa Research, Palo Alto CA 94306, USA\\ \email{$\{$miyeh, zzhuang, wzhan, liawang$\}$@visa.com}}}
\maketitle
\setcounter{footnote}{0}

\begin{abstract}
The multivariate time series generated from merchant transaction history can provide critical insights for payment processing companies.
The capability of predicting merchants' future is crucial for fraud detection and recommendation systems.
Conventionally, this problem is formulated to predict one multivariate time series under the \textit{multi-horizon} setting.
However, real-world applications often require more than one future trend prediction considering the uncertainties, where more than one multivariate time series needs to be predicted.
This problem is called \textit{multi-future prediction}.
In this work, we combine the two research directions and propose to study this new problem: \textit{multi-future}, \textit{multi-horizon} and \textit{multivariate} time series prediction.
This problem is crucial as it has broad use cases in the financial industry to reduce the risk while improving user experience by providing alternative futures.
This problem is also challenging as now we not only need to capture the patterns and insights from the past but also train a model that has a strong inference capability to project multiple possible outcomes.
To solve this problem, we propose a new model using convolutional neural networks and a simple yet effective encoder-decoder structure to learn the time series pattern from multiple perspectives.
We use experiments on real-world merchant transaction data to demonstrate the effectiveness of our proposed model.
We also provide extensive discussions on different model design choices in our experimental section.
\end{abstract}

\section{Introduction}
\label{sec-introduction}
The advances in digital payment systems in recent years have enabled billions of payment transactions to be processed every second.
Merchant transaction history is a prevalent data type of payment processing systems.
From airlines to book stores, the aggregation process for several critical features (e.g., the total amount of money spent in a book store between 5-6 pm) happens on an hourly-base, and the patterns of the transactions are monitored.
Monitoring such patterns is essential to applications such as fraud detection (i.e., by observing the deviation from the \textit{regular} trends) and shopping recommendation (i.e., by observing similar transaction histories).

One crucial step to build these applications is \textit{estimating} every merchant's \textit{future}, where each feature is predicted hourly in a ``rolling'' fashion.
However, the constant high computation cost makes such an approach unrealistic in real-world scenarios.
Instead, multi-horizon prediction, where the goal is to predict multiple time steps \textit{at one time} instead of only one time step in the future, is a preferred approach in this scenario.
\begin{figure}[t]
    \centering
    \includegraphics[width=0.6\linewidth]{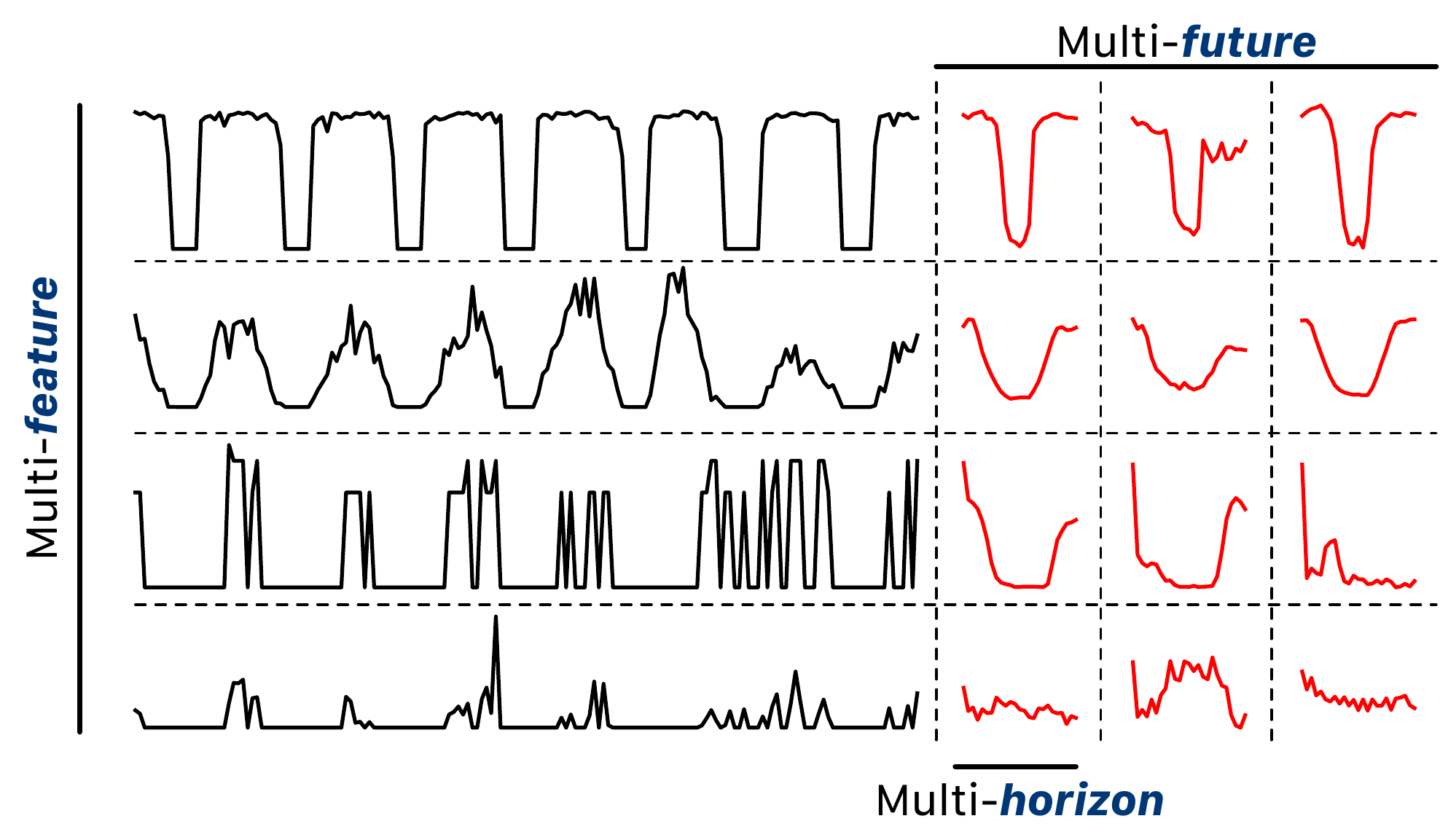}
    \caption{Multi-\textit{future} Multi-\textit{horizon} Multi\textit{variate} Time Series Prediction. }
    \label{fig.problem}
\end{figure}

Moreover, predicting only one \textit{future} in the financial industry may not be ideal as multiple factors can have a significant impact on the volume of merchant transactions.
For example, severe weather may or may not have a significant impact on restaurants' business.
If a model only predicts one trend signaling lower transaction volume, and there are still a large number of orders, it may cause a higher false-positive rate of credit card declines.
Therefore, predicting multiple possible features is a more realistic approach.
This problem is called \textit{multi-future time series prediction}.

We depict the overall scope of this problem in Fig.~\ref{fig.problem}.
For instance, the first feature in Fig.~\ref{fig.problem} is the per hour transaction volume.
The first and third possible future shows the business remains a similar trend as the previous history, while the second possible future shows jitter patterns.
By knowing these two alternative patterns for the future transaction volume, downstream systems could prepare alternative plans for each of the possible futures.

Time series prediction is a well-studied problem~\cite{box1970time}.
Recent works on time series prediction~\cite{taieb2015bias,wen2017multi} take a new look at the multi-horizon time series prediction problem from the neural network perspective.
Specifically, Taieb and Atiya~\cite{taieb2015bias} proposes to train a neural network with a target consisting of multi-steps into the future; Wen \textit{et al.}~\cite{wen2017multi} approaches this problem from sequence-to-sequence perspective.
However, none of those above work deals with the multi-future prediction problem.
The closest problem formulation solved by prior work is described in~\cite{tang2019multiple}, where multiple possible future trajectories of vehicles are predicted, but the system proposed is designed specifically for modeling driving behaviors, which is fundamentally different from the multivariate time series in the financial industry.
Predicting multiple possible futures on multivariate time series in a multi-horizon setting remains challenging in the real-world applications as the real-world data may exhibit frequent patterns that discourage the prediction of various futures.


In this work, we study the problem of \textit{multi-future, multi-horizon multivariate time series prediction}.
To tackle this challenge, we propose a novel model with two sub-networks: a \textit{shape} sub-network responsible for learning the time series shape patterns, and a \textit{scale} sub-network responsible for learning the magnitude and offset of the time series.
Each sub-network is built using a simple yet effective encoder-decoder stack with layers (e.g., Linear Layers, Convolutional Neural Networks, Max/Average Pooling Layers) that have lower computation overhead considering the real-world application.
We summarize our contributions as follows:
\begin{itemize}
    \item We analyze the problem of multi-future prediction for multi-horizon multivariate time series data.
    \item We propose and design a novel architecture for learning both the shape and scale of the time series data.
    \item We conduct our experiments on real-world merchant transactions and demonstrate the effectiveness of our proposed model.
\end{itemize}


\section{Notations and Problem Definition}
\label{sec-problem}
In this section, we present several important definitions in this work.
The most fundamental definitions are \textit{time series} and \textit{multivariate time series}.

\begin{definition}[Time Series]
    {\rm
    A \textit{time series} is an ordered set of real-valued numbers.
    For a time series~$\boldsymbol{\tau}$ of length~$n$, $\boldsymbol{\tau}$ is defined as $[t_1, \cdots, t_n] \in \mathbb{R}^n$.
    }
    \label{def-ts}
\end{definition}
\begin{table}[t]
\centering
\caption{Important Notations.}
\begin{tabular}{c|p{3.5in}}
\toprule
Notation & Meaning \\ \hline
 $\pmb{\tau} = [t_1, \cdots, t_n]$ & Time series of length $n$. \\
 $\mathbf{T} = [\pmb{\tau}_1, \cdots, \pmb{\tau}_d]$ & Multivariate time series.  \\
 $\widehat{\mathbb{T}} = \left [ {\widehat{\mathbf{T}}_1}, \cdots, {\widehat{\mathbf{T}}_f} \right ]$ &  Set of $f$ \textit{predicted} time series. \\
 $\mathbf{T}[i]$ & The $i$-th time step of the multivariate time series $\mathbf{T}$. \\
 $\mathbf{T}[i: j]$ & The multivariate time series between time $i$ and $j$ of the multivariate time series $\mathbf{T}$. \\
 $\bar{\mathbf{T}}$ & Multivariate time series ground truth. \\
 $\mathsf{M}$ & Multi-future time series prediction model.  \\
 $\pmb{h}$ & Feature vector output by a encoder. \\
 $\mathbf{S}$ & Shape bank (a matrix). \\
 $\pmb{r}$ & Activation vector. \\
 $\widehat{\pmb{\alpha}}^{\text{(sp)}}_{i,j}$ & The shape prediction for $i$-th future for the $j$-th feature. \\
 $\widehat{\boldsymbol{A}}^{\mathrm{(sp)}}_{i}$ & The multivariate shape prediction for the $i$-th future. \\
 $\mu_{i, j}$ & The offset of the $i$-th future, $j$-th feature. \\
 $\sigma_{i, j}$ & The magnitude of the $i$-th future, $j$-th feature. \\
 $i_\text{oc}$ & Oracle future index. \\
 \bottomrule
\end{tabular}
\end{table}

\begin{definition}[Multivariate Time Series]
    {\rm
    A \textit{multivariate time series} is a set of co-evolving time series, denoted as $\mathbf{T}$.
    For a multivariate time series~$\mathbf{T}$ of length~$n$ with $d$~features, $\mathbf{T}$ is defined as $[\boldsymbol{\tau}_1, \cdots, \boldsymbol{\tau}_{d}] \in \mathbb{R}^{n \times d}$, where each $\boldsymbol{\tau}_i \in \mathbf{T}$ denotes a co-evolving time series in the set.
    }
\end{definition}

Moreover, given a multivariate time series $\mathbf{T}$, we use $\mathbf{T}[i]$ to denote the values at time~$i$, and $\mathbf{T}[i:j]$ to denote the values of the multivariate time series from time~$i$ to time~$j$.

With the basic notation defined, we are ready to define the problem we are solving in this work: \textit{multi-future, multi-horizon, multivariate} time series prediction

\begin{definition}[Multi-future Time Series Set]
    {\rm
    A multi-future time series set $\widehat{\mathbb{T}}$ is a set of $f$ \textit{predicted} multivariate time series:
    \begin{equation}
        \widehat{\mathbb{T}} = \left [ {\widehat{\mathbf{T}}_1}, \cdots, {\widehat{\mathbf{T}}_f} \right ]
    \end{equation}
    where each multivariate time series~$\widehat{\mathbf{T}}_i \in \widehat{\mathbb{T}}$ is the prediction of a possible future.
    }
\end{definition}

\begin{definition}[Multi-future Time Series Prediction Model]
    {\rm
    Given the current time index as $i$, a multi-future time series prediction model~$\mathsf{M}$ predicts~$f$ possible futures for the next $n_h$ time horizon using the multivariate time series~$\textbf{T}$ from the past $n_p$ time points:
    \begin{equation}
        \widehat{\mathbb{T}}\gets \mathsf{M}\left(\mathbf{T}\left [i - n_p + 1:i\right ]\right)
    \end{equation}
    }
\end{definition}

\begin{definition}
    {\rm
    The goal of \textit{multi-future multi-horizon multivariate time series prediction problem} is to train a multi-future time series prediction model~$\mathsf{M}$ that minimizes the following performance measurement:
    \begin{equation}
    \sum_i \min_{1 \leq j \leq f} \texttt{error}(\mathbf{\widehat{T}}_{j}, \mathbf{T}[i+1:i+n_h])
    \end{equation}
    where $\texttt{error}(\cdot)$ is a function computing error-based performance measurement.
    One possible choice for $\texttt{error}(\cdot)$ is Root Mean Squared Error (\texttt{RMSE}).
    }
\end{definition}


\section{Model Architecture}
\label{sec-shapenet}
\begin{figure}[t]
    \centering
    \includegraphics[width=0.8\linewidth]{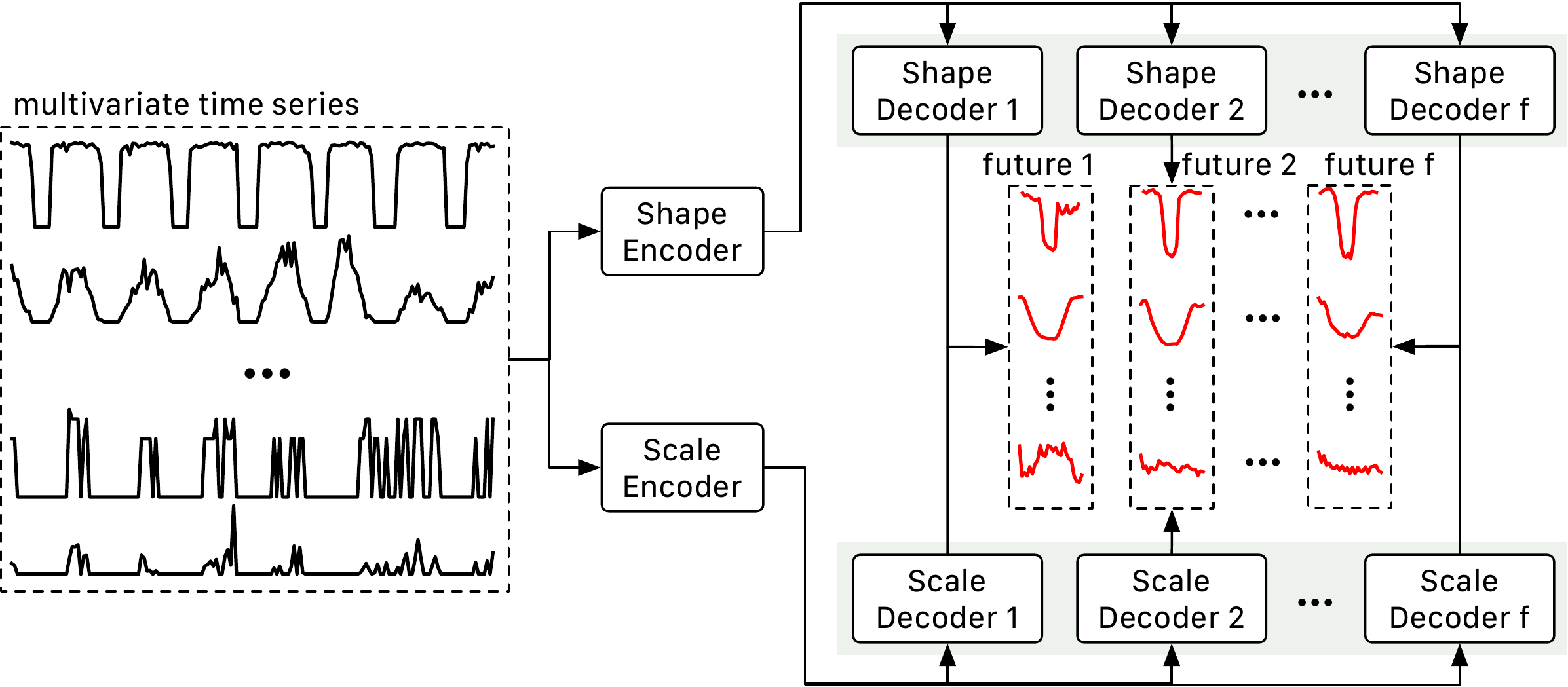}
    \caption{The proposed model. }
    \label{fig.model}
\end{figure}
We use Fig.~\ref{fig.model} to illustrate the overall architecture of the proposed model.
The proposed model has two sub-networks: a \textbf{shape} sub-network and a \textbf{scale} sub-network. Both sub-networks use the input multivariate time series~$\mathbf{T}$ in \textit{parallel}; then, the output of each sub-network is combined to form the prediction.

Each sub-network has an encoder and a set of decoders (i.e., an ensemble of decoders).
We refer to the encoder and decoders of the shape sub-network as \textbf{shape encoder} and \textbf{shape decoder}.
Similarly, we refer to the encoder and decoders of the scale sub-network as \textbf{scale encoder} and \textbf{scale decoder}.

While there is only one encoder, the number of decoders in each ensemble corresponds to the number of possible futures to predict. Specifically, each decoder in each ensemble is associated with a possible future.
For example, shape decoder~$i$ and scale decoder~$i$ are both associated with the $i$-th future $\mathbf{\widehat{T}}_{i}$.

\subsection{Shape Sub-network}
\begin{figure}[t]
    \centering
    \includegraphics[width=0.7\linewidth]{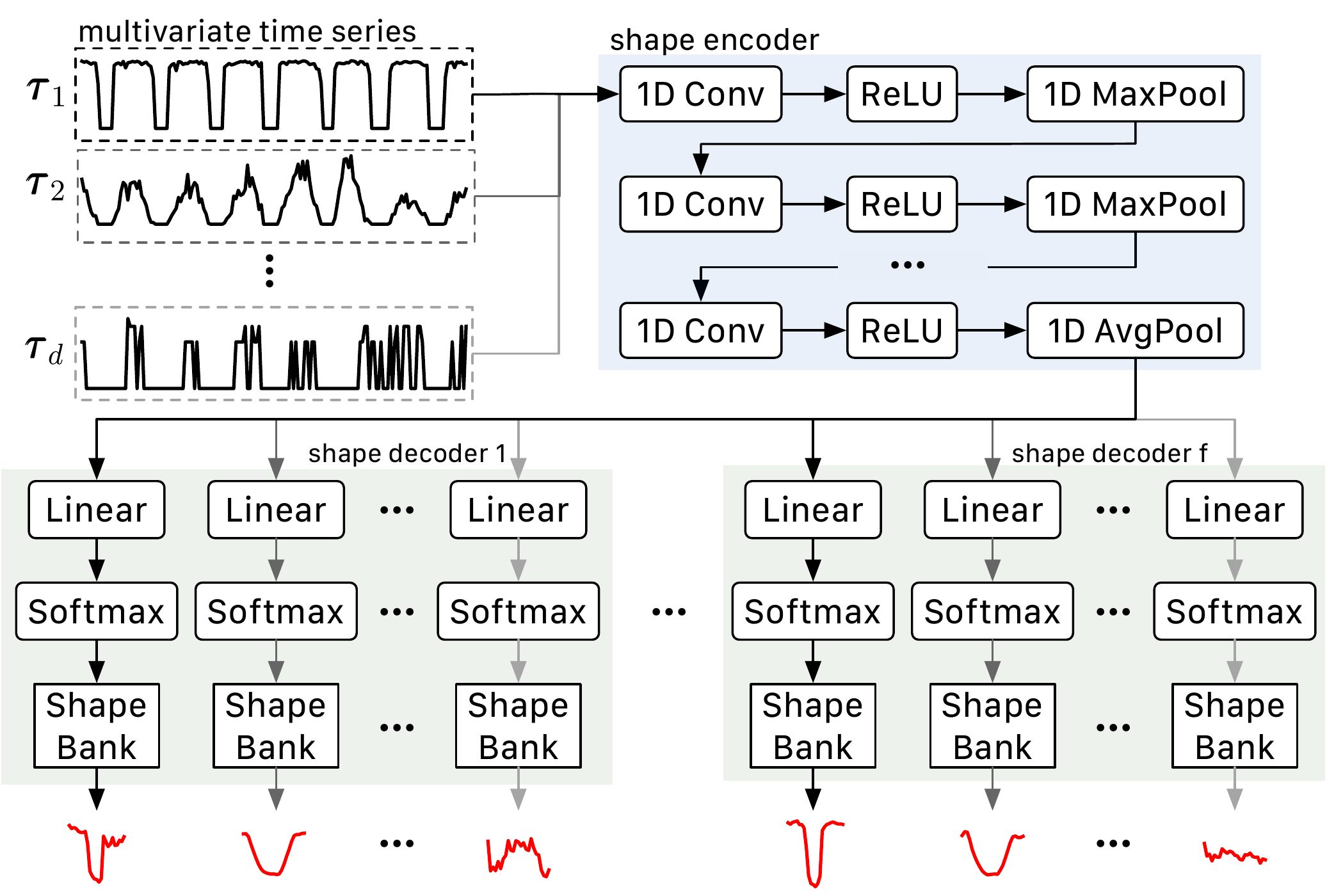}
    \caption{The shape sub-network contains a single encoder and a decoder ensemble.}
    \label{fig.shape_model}
\end{figure}

We illustrate our proposed \textbf{shape sub-network} in Fig.~\ref{fig.shape_model}.
The shape sub-network consists of a single shape encoder and an ensemble of shape decoders.
The purpose of the shape encoder is to capture relevant information for synthesizing the shape of the predicted time series.

\subsubsection{Encoder.} The encoder is built by stacking 1D convolution layer (denoted as \texttt{Conv}), rectified linear unit activation function (denoted as \texttt{ReLU}), 1D max pooling layers (denoted as \texttt{MaxPool}), and a 1D average pooling layer (denoted as \texttt{AvgPool}). Each stack is composed of a \texttt{Conv}, a \texttt{ReLU} and a \texttt{MaxPool} or \texttt{AvgPool}. That is, we use \texttt{AvgPool} instead of \texttt{MaxPool} in the last block to summarize the input time series along the temporal direction.
For an encoder with $l$ layers, the encoder can be expressed as:
\begin{align*}
    \pmb{h}_1 &= \texttt{MaxPool}\left(\texttt{ReLU}\left(\texttt{Conv}\left(\mathbf{T}\right)\right)\right)\\
    & \hdots \\
    \pmb{h} &= \texttt{AvgPool}\left(\texttt{ReLU}\left(\texttt{Conv}\left(\pmb{h}_{l-1}\right)\right)\right)
\end{align*}

The encoder processes the input multivariate time series~$\mathbf{T} \in \mathbb{R}^{n_p \times d}$ into a fixed-size vector representation $\pmb{h} \in \mathbb{R}^{d_h}$, where $d_h$ is the dimension of the representation.
In our particular implementation, for all \texttt{Conv} layers, the receptive field size is set to~3, and the number of channels is set to~64.
For the all \texttt{MaxPool}, the window size and stride are both set to~2.

Since our particular parameter settings for the layers would reduce the length of the time series by half each time the input passing trough a block, there are $\lfloor\log_2{n_p}\rfloor$ blocks being used to process the input time series.
Because $\log_2{n_p}$ is not guaranteed to be an integer, we use an adaptive pooling layer (or global pooling layer) for the last pooling layer (i.e., \texttt{AvgPool}) to make sure there is only one pooling window and the pooling window covers the whole intermediate output of the previous layer.

Given a multivariate time series~$\mathbf{T}$, our shape encoder would output the corresponding hidden representation vector~$\pmb{h}$ of size~64.

\subsubsection{Decoder.}
Each decoder uses a shallow design with two major components: a \textit{shape bank} and a \textit{softmax regression}.
In our model, each decoder is responsible for predicting one possible \textit{futures}.
For example, there will be three decoders if three different futures are to be predicted.

The shape bank stores a set of shape \textit{templates} for synthesizing the prediction's shape for a specific feature dimension. Thus, in every shape decoder, the number of shape banks is the same as the number of feature dimensions.
Moreover, each shape template has the same length as the output time series.
Given there are a total of $d$ feature dimensions in the multivariate time series, $n_h$ as the time horizon (i.e., length of the shape template) and a total of $n_s$ shape templates to be learned, each of the $d$~shape bank is implemented as a matrix of $\mathbb{R}^{n_s \times n_h}$.
Note, the shape banks can be either supplied by the user and/or refine/learned during the training process.

There are also $d$~\texttt{softmax} regression models, where each of them corresponds to one of the $d$~shape banks.
Each of the \texttt{softmax} regression models dictates how the templates stored in the corresponding shape bank are combined to form the shape prediction.
\begin{figure}[t]
    \centering
    \includegraphics[width=0.7\linewidth]{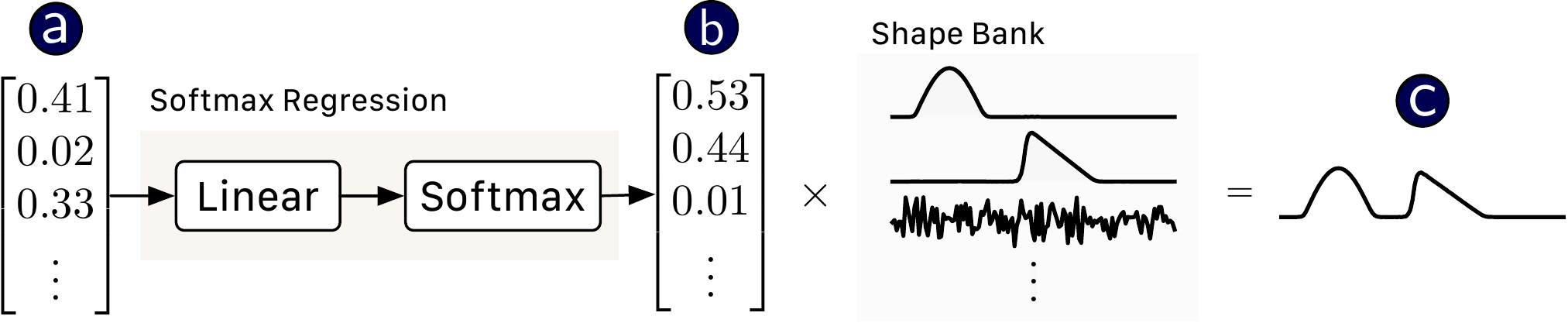}
    \caption{Synthesizing process for the shape decoder for a dimension.}
    \label{fig.shape_decoder}
\end{figure}

Given the hidden representation~$\pmb{h}$ produced by the shape encoder and respective shape bank $\mathbf{S} \in \mathbb{R}^{n_s \times n_h}$ with $n_s$ templates for the $j$-th feature dimension, the respective \texttt{softmax} regression model for the $j$-th feature dimension outputs the activation vector~$\pmb{r} \in \mathbb{R}^{n_s}$ where each value $r_k \in \pmb{r}$ is the weighting for each shape template $\mathbf{s}_k \in \mathbf{S}$.
Thereafter, the shape prediction $\widehat{\boldsymbol{\alpha}}^{\mathrm{(sp)}}_{j}$ for the $j$-th feature dimension is computed as follows:
\begin{equation}
    \widehat{\boldsymbol{\alpha}}_{j}^{(\text{sp})} = \pmb{r}\mathbf{S}
\end{equation}
Since $\widehat{\boldsymbol{\alpha}}^{\mathrm{(sp)}}_{j}$ represents only one future for the $j$-th feature dimension, we further denote $\widehat{\boldsymbol{\alpha}}^{\mathrm{(sp)}}_{i,j}$ as the prediction of the $i$-th future for the $j$-th feature dimension.
To simplify the notation in later section, we use $\widehat{\boldsymbol{A}}^{\mathrm{(sp)}}_{i} \in \mathbb{R}^{d \times n_h}$ to denote the multivariate shape prediction for the $i$-th future.
Fig.~\ref{fig.shape_decoder} shows an example of a forward pass for synthesizing the shape prediction for one of the dimensions where \circled{a} is the hidden representation output by the shape encoder, \circled{b} is the activation vector computed by the \texttt{softmax} regression model, and \circled{c} is the shape prediction.
The example presented in Fig.~\ref{fig.shape_decoder} shows another benefit of our shape decoder design: the shape decoder is an interpretable model.
By showing the activation vector and the associated shape template to the user, the user can understand both the synthesizing process and the relationship between the input time series and the relevant shape template in the shape bank.

\subsection{Scale Sub-network}

\begin{figure}[t]
    \centering
    \includegraphics[width=0.75\linewidth]{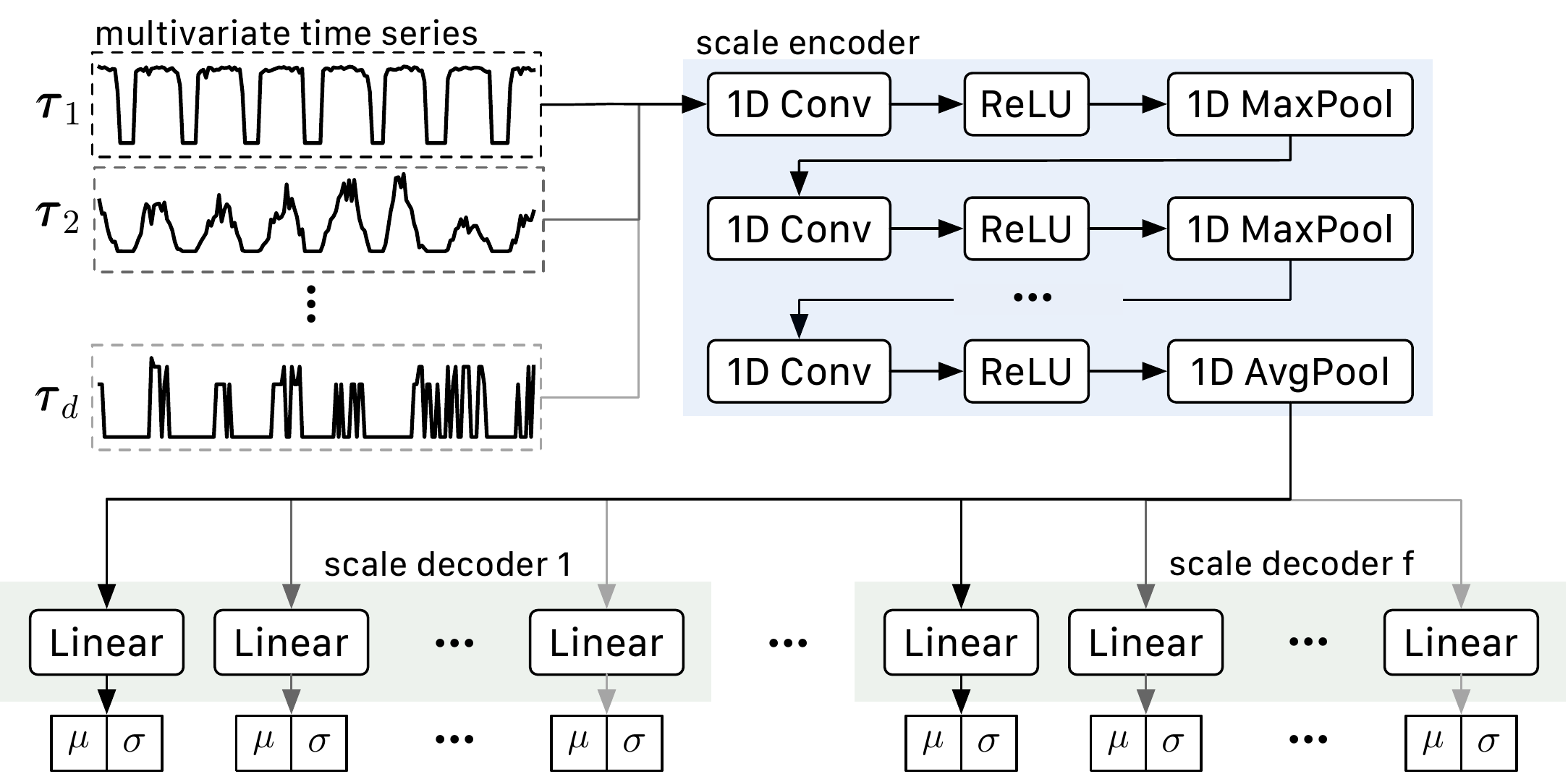}
    \caption{The scale sub-network contains a single encoder and a decoder ensemble.}
    \label{fig.scale_model}
\end{figure}

The scale sub-network (see Fig.~\ref{fig.scale_model}) has a similar high-level encoder-decoder structure as the shape sub-network.
However, the objective here is different from the shape sub-network.
The purpose of the scale sub-network is to capture relevant information for synthesizing the scale of the predicted time series.

\subsubsection{Encoder.}
The scale encoder has the same \texttt{1D Conv} stack as the shape encoder, and we use the same hyper-parameter settings for the \texttt{1D Conv} layers.
Therefore, given a multivariate time series~$\mathbf{T}$, the scale encoder also outputs a hidden representation vector~$\pmb{h}$ of size~64.
Depending on the data, it is possible to have a simplified model with fewer trainable parameters for the scale encoder as the scale information could be relatively simpler comparing to the shape information.
However, using the same architecture for the shape encoder and the scale encoder has proven the most effective on our merchant transaction data.
Searching for a better encoder architecture for both the scale and the shape encoders is an interesting future direction.

\subsubsection{Decoder.}
For the scale decoder, we also use a shallow model for the prediction process, which is similar to the shape decoder.
However, instead of the \texttt{softmax} regression-based model, we use a linear model to predict the scale (i.e., offset~$\mu$ and magnitude~$\sigma$) of the output time series for each dimension.
Similar to the notation we used for the shape prediction, we use $\mu_{i,j}$ and $\sigma_{i,j}$ to denote the $\mu$ and $\sigma$ for $i$-th future and $j$-th feature dimension.

To form the $i$-th future prediction~$\widehat{\mathbf{T}}_{i}$, we combine scale sub-network's output (i.e., $\mu_{i,j}$ and $\sigma_{i,j}$) with shape sub-network's output (i.e., $\widehat{\boldsymbol{\alpha}}_{i,j}^{(\text{sp})}$) using the following equation:
\begin{equation}
    \widehat{\mathbf{T}}_{i} =
    \begin{bmatrix}
    \mu_{i,1} \widehat{\boldsymbol{\alpha}}^{(\text{sp})}_{i,1} + \sigma_{i,1}\\
    \vdots\\
    \mu_{i,d} \widehat{\boldsymbol{\alpha}}^{(\text{sp})}_{i,d} + \sigma_{i,d}\\
    \end{bmatrix}
\end{equation}

\subsubsection*{Can we estimate the probability of each possible future?}
Such a problem can be solved with the concept of expert classifier~\cite{babu2018divide}.
An expert classifier is a model that predicts the most likely outcome out of all of the possible futures returned by the main model given an input.
We can use an architecture similar to the shape encoder for expert classifiers by adding an additional linear and softmax layer to predict the most likely outcome.
The expert classifier can be optimized by 1) feeding an input through the main network and stored the best-fitted outcome id as the ground truth label and 2) use the cross-entropy loss with the ground truth label and the input to optimize the expert classifier.

\section{Training Algorithm}
\label{sec-training}

To ensure that each sub-network captures the corresponding information under multi-future prediction setting, we use the loss function shown in Eq.~\ref{eq-loss}:
\begin{equation}
\label{eq-loss}
\mathcal{L}(\mathbf{T}, \mathbf{\bar{T}}) = \texttt{RMSE}(\mathbf{\bar{T}}, \mathbf{\widehat{T}}_{i_\text{oc}}) + \gamma \texttt{NRMSE}(\mathbf{\bar{T}}, \widehat{\boldsymbol{A}}^{\mathrm{(sp)}}_{i_\text{oc}})
\end{equation}
where $\mathbf{T} \in \mathbb{R}^{n_p \times d}$ is the input multivariate time series, $\mathbf{\bar{T}} \in \mathbb{R}^{n_h \times d}$ is the ground truth time series (i.e., the multivariate time series for the next $n_h$ time steps), $\texttt{RMSE}(\cdot)$ is a function computes Root Mean Squared Error, $\texttt{NRMSE}(\cdot)$ is a function computes the Normalized Root Mean Square Error, $\mathbf{\widehat{T}}_{i_\text{oc}}$ is the prediction for $i_\text{oc}$-th future, $\widehat{\boldsymbol{A}}^{\mathrm{(sp)}}_{i_\text{oc}}$ is the shape prediction for $i_\text{oc}$-th future, $i_\text{oc}$ is the oracle future index and $\gamma \in \mathbb{R}$ is a hyper-parameter balancing the $\texttt{RMSE}(\cdot)$ and $\texttt{NRMSE}(\cdot)$ terms.
We set $\gamma$ to 1 in all our experiments.
The oracle future index~$i_\text{oc}$ is computed using Eq.~\ref{eq-ioc}, and it is only determined based on the shape prediction.
\begin{equation}
\label{eq-ioc}
i_\text{oc} = \argmin_{1 \leq j \leq f} \texttt{NRMSE}(\mathbf{\bar{T}}, \widehat{\boldsymbol{A}}^{\mathrm{(sp)}}_{j})
\end{equation}

The loss is computed by aggregating the \texttt{RMSE} and the \texttt{NRMSE} between the best-predicted future and the ground truth.
Only the shape prediction determines the best-predicted future because shape prediction is the harder problem comparing to the scale prediction problem.
Specifically, the $\texttt{NRMSE}(\mathbf{\bar{T}}, \widehat{\boldsymbol{A}}^{\mathrm{(sp)}}_{i_\text{oc}})$ term is computed by first z-normalized~\cite{rakthanmanon2012searching} each dimension of the ground truth~$\mathbf{\bar{T}}$, then compute the \texttt{RMSE} between the normalized ground truth and the shape prediction $\widehat{\boldsymbol{A}}^{\mathrm{(sp)}}_{i_\text{oc}}$.
Because both $\texttt{RMSE}(\mathbf{\bar{T}}, \mathbf{\widehat{T}}_{i_oc})$ and $\texttt{NRMSE}(\mathbf{\bar{T}}, \widehat{\boldsymbol{A}}^{\mathrm{(sp)}}_{i_\text{oc}})$ are computed using the best possible future in the set of multiple predicted future, we also refer to them as oracle \texttt{RMSE} and oracle \texttt{NRMSE}; both are special cases of the oracle loss function~\cite{guzman2012multiple,lee2016stochastic}.

We use Algorithm~\ref{alg-train} to training the model.
The main input to the algorithm is the training data~$\mathbf{T} \in \mathbb{R}^{n \times d}$, and the outputs are the trained model~$\mathsf{M}$.
First, $\mathsf{M}$ is initialized in \texttt{Line 2}.
The main training loop starts at \texttt{Line 3}.
At the beginning of each iteration, a mini-batch consist of the input $\mathbf{T}_\text{batch}$ and the ground truth $ \mathbf{\bar{T}}_\text{batch}$ is sampled from~$\mathbf{T}$ as shown in \texttt{Line 4}.
The input~$\mathbf{T}_\text{batch}$ is a tensor of size $(n_b, n_p, d)$, and the corresponding ground truth~$\mathbf{\bar{T}}_\text{batch}$ is a tensor of size $(n_b, n_h, d)$ where $n_b$ is the batch size.
For example, if $\mathbf{T}_\text{batch}[i, :, :]$ is sampled from $\mathbf{T}[j+1:j + n_p, :]$, $\mathbf{\bar{T}}_\text{batch}[i, :, :]$ will contain $\mathbf{T}[j + n_p + 1:j + n_p + n_h, :]$.
Next, from \texttt{Line 5} to \texttt{Line 8}, the loss for each instance in the mini-batch is computed using Eq.~\ref{eq-loss} and Eq.~\ref{eq-ioc}.
The oracle future index is determined using Eq.~\ref{eq-ioc} in \texttt{Line 7} and the loss is computed using Eq.~\ref{eq-loss} in \texttt{Line 8}.
The loss for each instance within the mini-batch is aggregated together.
Last, in the iteration, the model~$\mathsf{M}$ is updated using the gradient computed using the loss at \texttt{Line 8}.
The trained model~$\mathsf{M}$ is returned at \texttt{Line 9} after the model is converged.
\begin{algorithm}[htb]
    \centering
    \caption{The Training Algorithm\label{alg-train}}
    \begin{algorithmic}[1]
        \Input{multivariate time series $\mathbf{T}$}
        \Output{model $\mathsf{M}$}
        \Function{train}{$\mathbf{T}$}
        \State $\textrm{InitializeModel}\left(\mathsf{M}\right)$
        \For{$i \gets 0 \textrm{ \textbf{to} } n_\text{iter}$}
        \State $\mathbf{T}_\mathrm{batch}, \mathbf{\bar{T}}_\mathrm{batch} \gets \textrm{ GetMiniBatch}(\mathbf{T}, n_p, n_h)$
        \State $loss \gets 0$
        \For{$\textrm{\textbf{each} } \mathbf{T}_{i}, \mathbf{\bar{T}}_{i} \in \mathbf{T}_\mathrm{batch}, \mathbf{\bar{T}}_\mathrm{batch}$}
        \State $i_\text{oc} \gets \textrm{ GetOracleFutureIndex}(\mathbf{T}_{i}, \mathbf{\bar{T}}_{i}, \mathsf{M})$ \Comment{Use Eq.~\ref{eq-ioc}.}
        \State $loss \gets \text{ComputeLoss}\left(\mathbf{T}_\mathrm{i}, \bar{\mathbf{T}}_\mathrm{i}, \mathsf{M}, i_\text{oc}\right)$ \Comment{Use Eq.~\ref{eq-loss}.}
        \EndFor
        \State $\mathsf{M} \gets \text{UpdateModel}(\mathsf{M}, loss)$
        \EndFor
        \State \Return{$\mathsf{M}$}
        \EndFunction
    \end{algorithmic}
\end{algorithm}

\section{Experiments}
\label{sec-experiments}
In this section, we aim to demonstrate the effectiveness of the proposed method by comparing it to the alternatives under both single-future prediction setup and multi-future prediction setup.
For single-future experiments, we use \texttt{RMSE} and \texttt{NRMSE} as the performance measurement.
For multi-future experiments, we use oracle \texttt{RMSE} and oracle \texttt{NRMSE} as the performance measurement.
While \texttt{RMSE} gives us the measure of the deviation of prediction from the ground truth in raw value, it does not measure the deviation in terms of shape.
To give us a complete picture of the proposed method's ability in prediction, we also choose to include \texttt{NRMSE} to differentiate different method's ability to predict the correct shape.
In the financial industry, both the raw values and the shape of the future trend are essential for decision making.
All the deep learning-based methods are implemented in PyTorch, and we use Adam optimizer~\cite{kingma2014adam} with the default parameters setting for optimization.

\subsection{Description of the Datasets}
\label{sec-exp-dataset}

We have organized four different datasets where each dataset consists of merchants from one of the following categories: department store, restaurants, sports facility, and medical services, denoted as Cat.1 $\sim$ Cat.4, respectively.

For each category, we randomly select 2,000 merchants located within California, United States.
The time series datasets consist of four features, and each is produced by computing the hourly aggregation of the following statistics: number of approved transactions, number of unique cards, a sum of the transaction amount, and rate of the approved transaction.
The training data consists of time series data from November 1, 2018, to November 23, 2018; the test data consists of a time series from November 24, 2018, to November 30, 2018.

As mentioned in Section~\ref{sec-introduction}, the goal of the system is to predict the next 24 hours given the last 168 hours (i.e., seven days).
We predict every 24 hours in the test data by supplying the latest 168 hours to the system.
For example, the transaction data of 168 hours in \texttt{Week-10} is used to predict the values of 24 hours on the Monday of \texttt{Week-11}.



\subsection{Evaluation of Architecture Design Choice}
\label{sec-exp-design}
In this section, we focus on evaluate the architecture design of our model.
As design choice explored in the above questions is agnostic to the number of future to predict, we evaluate both the designs implemented in the proposed model architecture and the alternatives under a single-future prediction setting.
Specifically, our experiments focus on answering the following questions:

\begin{figure}[t]
    \centering
    \includegraphics[width=0.8\textwidth]{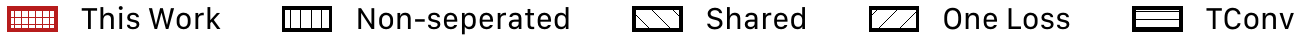}
\begin{minipage}{.49\textwidth}
\centering
    \begin{subfigure}[t]{0.49\textwidth}
        \centering
        \includegraphics[page=1, width=\textwidth]{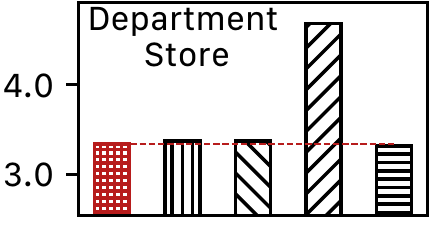}
    \end{subfigure}
    \begin{subfigure}[t]{0.49\textwidth}
        \centering
        \includegraphics[page=2, width=\textwidth]{figexp_rmse_shapenet_others.pdf}
    \end{subfigure}

    \begin{subfigure}[t]{0.49\textwidth}
        \centering
        \includegraphics[page=3, width=\textwidth]{figexp_rmse_shapenet_others.pdf}
    \end{subfigure}
    \begin{subfigure}[t]{0.49\textwidth}
        \centering
        \includegraphics[page=4, width=\textwidth]{figexp_rmse_shapenet_others.pdf}
    \end{subfigure}
    \caption{Performance results in \texttt{RMSE}.}
    \label{figexp_rmse}
\end{minipage}
\begin{minipage}{.49\textwidth}
\centering
    \begin{subfigure}[t]{0.49\textwidth}
        \centering
        \includegraphics[page=1, width=\textwidth]{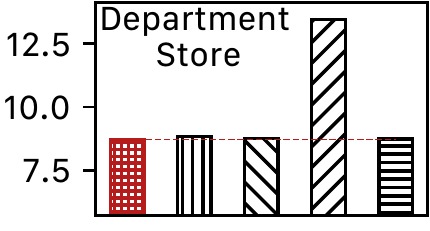}
    \end{subfigure}
    \begin{subfigure}[t]{0.49\textwidth}
        \centering
        \includegraphics[page=2, width=\textwidth]{figexp_nrmse_shapenet_others.pdf}
    \end{subfigure}

    \begin{subfigure}[t]{0.49\textwidth}
        \centering
        \includegraphics[page=3, width=\textwidth]{figexp_nrmse_shapenet_others.pdf}
    \end{subfigure}
    \begin{subfigure}[t]{0.49\textwidth}
        \centering
        \includegraphics[page=4, width=\textwidth]{figexp_nrmse_shapenet_others.pdf}
    \end{subfigure}
    \caption{Performance results in \texttt{NRMSE}.}
    \label{figexp_nrmse}
\end{minipage}
\end{figure}

\subsubsection{Does it benefit from having dedicated shape sub-network and scale sub-network?}
Instead of using dedicated encoder-decoder sub-network to model different aspects of the time series data (i.e., shape and scale), one could simply just using one encoder-decoder to model time series data.
To evaluate the effectiveness of the dual sub-network design, we compare the proposed model to an alternative model where it just consists of an encoder with the shape-encoder architecture and a decoder with the shape-decoder architecture.
As the alternative does not use separate sub-networks, we refer to this alternative as \textit{non-separated model}.
In Fig.~\ref{figexp_rmse} and Fig.~\ref{figexp_nrmse}, the performance of the proposed method (i.e., \iconthiswork) and the alternative (i.e., \iconnosep) are presented.
The proposed method has a higher or equal performance comparing to a non-separated model in all datasets with both \texttt{RMSE} and \texttt{NRMSE}, especially when the performance is measured with \texttt{NRMSE}.

\subsubsection{Can we share the encoder in shape/scale sub-network?}
As demonstrated by the experiments answering the last question, it is beneficial to have dedicated sub-network for modeling different aspects of time series data.
However, does it require a dedicated encoder for the proposed model to function effectively, or is it possible to share the encoder for both shape and scale sub-network?
To answer this question, we have implemented an alternative model structure where there is only one encoder, and the output of the encoder $h_l$ is feed to a dedicated shape decoder and scale decoder.
Because the alternative model shared the encoder, we call this alternative model \textit{shared model}.
The performance for the proposed model (i.e., \iconthiswork) and the shared model (i.e., \iconshared) is shown in Fig.~\ref{figexp_rmse} and Fig.~\ref{figexp_nrmse}.
The overall conclusion is similar to the last question, and the proposed method has a higher or equal performance comparing to the alternative in all datasets with both performance measures.

\subsubsection{Does it benefit to have both \texttt{RMSE} and \texttt{NRMSE} terms in the loss instead of just \texttt{RMSE}?}
Another innovation of our model design is the loss function: we use both \texttt{RMSE} and \texttt{NRMSE} in the loss function to guide the sub-networks to learn the corresponding aspect of the time series data.
In other words, the loss function design pushes the shape sub-network to model the shape of the time series and the scale sub-network to model the scale of the time series.
To test this hypothesis, we use an alternative loss to train the same model where the alternative loss only consists of the \texttt{RMSE} term.
Since this alternative only uses one \texttt{RMSE} in the loss, we call this alternative method \textit{one loss}.
In Fig.~\ref{figexp_rmse} and Fig.~\ref{figexp_nrmse}, we use \iconthiswork to denote the purposed model train using the loss function with both \texttt{RMSE} and \texttt{NRMSE}; we use \icononeloss to denote the method trained only with \texttt{RMSE} loss.
The design of the loss has a noticeable impact on the model's performance, the amount of improvement obtained by using the proposed loss function ranging from 15\% to 59\% compared to the alternative.
It is crucial to use the proposed loss function when training model with the proposed architecture.

\subsubsection{How does the shallow shape bank design comparing to the commonly seen deep transposed convolutional network?}
One alternative design for the decoder is the transposed convolutional network~\cite{radford2015unsupervised}, which is relatively deep comparing to the shape bank design adopted in the proposed model.
To compare our shallow design with the transposed convolutional network design, we implement a transposed convolutional (\texttt{TConv}) shape decoder as follows.
Given an activation vector~$\pmb{r}$, the shape prediction $\widehat{\boldsymbol{\alpha}}^{\mathrm{(sp)}}_{j}$ for the $j$-th feature dimension is computed by:
\begin{align*}
    \pmb{h}_0 &= \texttt{ReLU}\left(\texttt{Linear}\left(\pmb{r}\right)\right)\\[-2pt]
    \pmb{h}_1 &= \texttt{Upsample}\left(\texttt{ReLU}\left(\texttt{TConv}\left(\pmb{h}_0\right)\right)\right)\\[-2pt]
    & \hdots \\[-2pt]
    \pmb{h}_l &= \texttt{Upsample}\left(\texttt{ReLU}\left(\texttt{TConv}\left(\pmb{h}_{l-1}\right)\right)\right)\\[-2pt]
    \widehat{\boldsymbol{\alpha}}^{\mathrm{(sp)}}_{j} &= \texttt{Conv}\left(\pmb{h}_l\right)
\end{align*}

Particularly, the \texttt{Linear} layer has 64 channels.
For both \texttt{TConv} and \texttt{Conv} layer, the receptive field size is set to 3, and the number of channels is set to 64.
Aside from the last \texttt{Upsample} layer, we use the upsampling factor of 2.
For the last \texttt{Upsample} layer, we set the output size to 24.
We set $l$ to 5 for generating a 24 sized time series.
The performance of the shallow shape bank design (i.e., \iconthiswork) and the deep \texttt{TConv} design (i.e., \icontcov) is shown in Fig.~\ref{figexp_rmse} and Fig.~\ref{figexp_nrmse}.
Aside from the department store dataset, the shape bank design achieves an improvement over the \texttt{TConv} design ranging from 14\% to 54\% for different datasets and performance measurements.
As different datasets consist of different patterns, the \texttt{TConv} decoder design only capable of synthesizing the patterns may appear in the department store dataset while struggle on synthesizing the patterns appears in other datasets.
Nevertheless, the proposed shape bank decoder has superb performance across the board comparing to the \texttt{TConv} decoder.

\newcolumntype{Y}{>{\centering\arraybackslash}X}
\begin{table}[t]
\centering
\caption{Comparison between the proposed method and the alternatives.}
\label{tab-alt_compare}
\centering
\begin{tabular}{ccccccccccc}
\toprule
\multirow{2}{*}{} & \multicolumn{5}{c}{\normalsize \texttt{RMSE}} & \multicolumn{5}{c}{\normalsize \texttt{NRMSE}} \\ \cmidrule(l){2-6} \cmidrule(l){7-11}
 & Cat.1 & Cat.2 & Cat.3 & Cat.4 & Avg. & Cat.1 & Cat.2 & Cat.3 & Cat.4 & Avg. \\ \midrule
\begin{tabular}[c]{@{}c@{}}Nearest \\[-2pt] Neighbor\end{tabular} & 5.9784 & 5.7083 & 8.6402 & 5.4327 & 6.4399 & 11.0063 & 6.8305 & 11.2740 & 9.5840 & 9.6737 \\ [5pt]
\begin{tabular}[c]{@{}c@{}}Linear \\[-2pt] Regression\end{tabular} & 3.6999 & \underline{\texttt{3.7094}} & 4.3324 & 3.7532 & 3.8737 & 10.5669 & 5.5393 & 10.2107 & 9.9300 & 9.0617 \\[5pt]
\begin{tabular}[c]{@{}c@{}}Random \\[-2pt] Forest\end{tabular} & 3.6918 & 4.2933 & 4.8370 & 3.9903 & 4.2031 & 8.9944 & 6.0096 & 10.5885 & 9.2405 & 8.7082 \\[5pt]
\begin{tabular}[c]{@{}c@{}}RNN \\[-2pt] (LSTM)\end{tabular} & 3.8138 & 3.7908 & 4.3121 & 3.3720 & 3.8222 & 9.9693 & 5.7273 & 10.2029 & 8.9879 & 8.7219 \\[5pt]
\begin{tabular}[c]{@{}c@{}}RNN \\[-2pt] (GRU)\end{tabular} & 3.8311 & 3.7381 & 4.3193 & 3.4631 & 3.8379 & 10.5512 & 5.6134 & 10.2579 & 9.1511 & 8.8934 \\[5pt]
\begin{tabular}[c]{@{}c@{}}This \\[-2pt] Work\end{tabular} & \underline{\texttt{3.3390}} & 3.7720 & \underline{\texttt{4.1736}} & \underline{\texttt{3.3530}} & \underline{\texttt{3.6594}} & \underline{\texttt{8.7137}} & \underline{\texttt{5.3977}} & \underline{\texttt{9.3612}} & \underline{\texttt{8.0980}} & \underline{\texttt{7.8927}} \\ \bottomrule
\end{tabular}
\end{table}

\subsubsection{How does the proposed architecture comparing to other methods?}
Besides of evaluating each design choice we mode for the proposed model, it is also important to compare the proposed model to both the baseline model and the state of the art model~\cite{wen2017multi} for multi-horizon time series prediction.
We compared the proposed model with the following alternative methods:
\begin{itemize}
    \item Nearest Neighbor.
    This method predicts the future by searching the nearest neighbor of the current time series from history.
    Once located the nearest neighbor of the training data, the following 24 hours of this nearest neighbor is used as the prediction~\cite{rakthanmanon2012searching}.
    \item Linear Regression.
    To apply this method to our problem, we first flatten the input multivariate time series into a feature vector, then we train 96 liner models using the flatten vector where each model is responsible for predicting one feature at one time step.
    The time series has four features, and we are predicting the next 24 time steps.
    We use $L2$ regularization when training the linear models.
    \item Random Forest.
    An ensemble method that trains a set of decision trees via bootstrap aggregating and random subspace method~\cite{breiman2001random,ho1995random}.
    Similar to the Linear Regression model, we flatten the input multivariate time series into a feature vector, and we formulate the multi-horizon prediction problem as a multi-output regression problem before applying off-the-shelf random forest implementation.
    \item Recurrent Neural Network.
    We use two different types of RNNs: Long Short-Term Memory (LSTM) and Gated Recurrent Unit (GRU).
    For each type of RNN, we use two layers to encode the input time series and a Multi-layer Perceptron (MLP) to predict the time series for the next 24 hours.
    This is the best performing architecture for multi-horizon time series prediction, according to~\cite{wen2017multi}.
\end{itemize}

The experiment result is summarized in Table~\ref{tab-alt_compare}.
Out of the three non-deep learning-based methods, linear regression has the best performance based on averaged \texttt{RMSE}.
When considering the averaged \texttt{NRMSE}, random forest outperforms the other two methods.
Although the random forest has worse \texttt{RMSE} comparing to linear regression, it is more capable of modeling the shape of time series.
When considering the alternative deep learning-based approaches, the RNNs achieve a superb \texttt{RMSE} comparing to linear regression and a similar \texttt{NRMSE} comparing to the random forest.
Such observation confirms that the model presented in ~\cite{wen2017multi} is also the best off-the-shelf method for modeling time series from transaction data.
Lastly, by comparing the proposed method to the alternatives, the proposed method demonstrates its superior performance in predicting both the raw values and the shape of the future time series.

\subsection{Evaluation of the Multi-future Learning Scheme}

\begin{figure}[t]
\centering
    \includegraphics[width=0.55\textwidth]{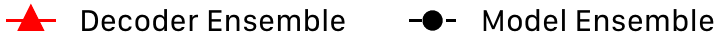}

    \begin{minipage}{.42\textwidth}
    \centering
        \begin{subfigure}[t]{0.65\textwidth}
            \centering
            \includegraphics[page=1, width=\textwidth]{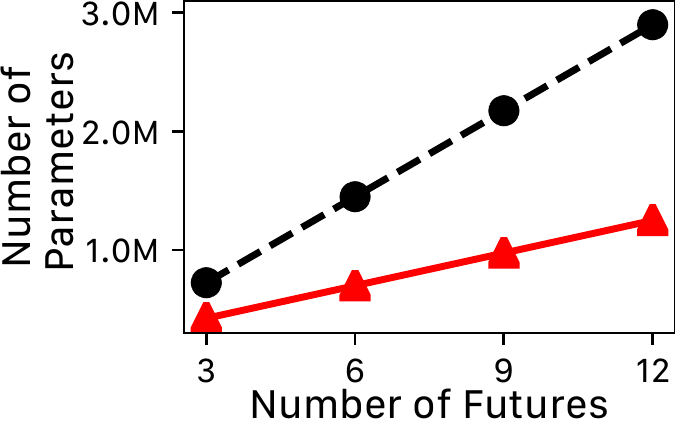}
            \label{figexp_perf_nparams}
        \end{subfigure}

        \begin{subfigure}[t]{0.65\textwidth}
            \centering
            \includegraphics[page=2, width=\textwidth]{figexp_perf.pdf}
            \label{figexp_perf_n_iter}
        \end{subfigure}
        \caption{Scalability Comparisons. }
        \label{figexp_scalability}
    \end{minipage}
    \begin{minipage}{0.55\textwidth}
    \centering
        \begin{subfigure}[t]{0.48\textwidth}
            \centering
            \includegraphics[page=3, width=\textwidth]{figexp_perf.pdf}
        \end{subfigure}
        \begin{subfigure}[t]{0.48\textwidth}
            \centering
            \includegraphics[page=4, width=\textwidth]{figexp_perf.pdf}
        \end{subfigure}
        \begin{subfigure}[t]{0.48\textwidth}
            \centering
            \includegraphics[page=5, width=\textwidth]{figexp_perf.pdf}
        \end{subfigure}
        \begin{subfigure}[t]{0.48\textwidth}
            \centering
            \includegraphics[page=6, width=\textwidth]{figexp_perf.pdf}
        \end{subfigure}
        \caption{Accuracy Comparisons.}
        \label{figexp_accuracy}
    \end{minipage}
\end{figure}

Traditionally, multi-future learning is achieved by training a deep learning ensemble where each model within the ensemble is responsible for making a possible prediction~\cite{lee2016stochastic,babu2018divide}.
In this section, we showcase the benefit of our multi-future learning scheme (i.e., decoder ensemble) comparing to the existing scheme (i.e., model ensemble)~\cite{lee2016stochastic,babu2018divide}.
We apply both multi-future learning schemes to the proposed model using the same training algorithm presented in Section~\ref{sec-training}.
We only consider the proposed model because the proposed model has superb performance comparing to the alternatives, as demonstrated in Section~\ref{sec-exp-design}.
To apply the model ensemble scheme to our model, we use multiple encoders, each corresponding to a decoder, to form the ensemble.
To showcase the difference in terms of the scalability and accuracy for each method, we present the comparison for the number of parameters, average run-time per iteration, and oracle \texttt{RMSE} for each of the dataset under a different number of futures settings in Fig.~\ref{figexp_scalability} and Fig.~\ref{figexp_accuracy}.
Regardless of the multi-future learning scheme, the number of parameters and training time grown linearly with the number of future (see Fig.~\ref{figexp_scalability}); the \texttt{RMSE}s of different datasets improves as the number of future (see Fig.~\ref{figexp_accuracy}).
When comparing the different multi-future learning schemes, the proposed model can achieve comparable \texttt{RMSE}s with 168\% and 443\% improvement in training time for 3 and 12 futures, respectively comparing to the existing scheme.
The number of parameters is only just 59\%, and 42\% relative to the model learned using an existing scheme when the number of futures is 3 and 12, respectively.
The \texttt{NRMSE} figures are omitted due to space limitation, but the conclusion remains the same.
In conclusion, the proposed multi-future learning scheme has better scalability comparing to the existing scheme with respect to the number of futures without sacrificing accuracy.
\section{Related Work}
\label{sec-related}
\noindent \textbf{Time Series Prediction}
is a well-studied problem dated back to~\cite{box1970time}.
Recent work utilizes the capability of neural networks to tackle this known problem:
Taieb and Atiya~\cite{taieb2015bias} train a neural network with multiple steps in the future as the target; Wen \textit{et al.}~\cite{wen2017multi} approach from the sequence-to-sequence learning perspective.
Another recent work~\cite{sen2019think} jointly trains a global matrix factorization model and a local temporal convolution network to model both global and local property of the time series.
Although the motivation for their model design is different from ours, their final design shares some similarities with our shape sub-network design.
The major difference between our model and the one proposed by Sen \textit{et al.}~\cite{sen2019think} is that they use a shared set of basic time series (i.e., shape bank) across all dimensions as they assume that all the dimensions have very similar behavior.
However, as shown in Fig.~\ref{fig.problem}, such an assumption does not hold for the time series generated from transaction records.
Our method also focuses on the separation between shape and scale information, which is not considered in the model presented in~\cite{sen2019think}.
Additional neural network-based methods are presented in~\cite{cerqueira2017arbitrated,de2018multivariate,shih2019temporal,fan2019multi} and~\cite{faloutsos2019forecasting} presents a comprehensive tutorial.
Nevertheless, none of these works attempt to produce a multi-future prediction for time series, and they only focus on predicting one determined future given a known set of time series.
Such fact limits the applicability of the methods mentioned above in real-world merchant transaction time series data.

\noindent \textbf{Multi-future Learning} or multiple-choice learning has attracted more attention recently.
Notable work includes:~\cite{guzman2012multiple}, where multiple hypotheses are generated for prediction tasks that incorporate user interactions or successive components;~\cite{lee2016stochastic} extended the learning algorithm presented in~\cite{guzman2012multiple} for deep learning models;~\cite{babu2018divide} uses a hierarchical incrementally growing CNN to count the number of people in a picture, the CNN grows following a binary tree structure throughout the training process, and an additional classifier is trained to route each test image through the network.
All the works mentioned above focus on computer vision applications; therefore, their methods cannot be directly applied to our problem.
In terms of problem formulation, the closest work to us is~\cite{tang2019multiple}; they are also solving time series prediction problems.
However, since their model is designed specifically for modeling vehicular behaviors and only focuses on predicting the very next time step, we could not directly apply their method to our problem.
In conclusion, these works are not tailored to tackle the challenges in time series prediction nor in multi-horizon multivariate time series prediction.
\section{Conclusion}
\label{sec-conclusion}
In this work, we identified the problem of multi-future multi-horizon prediction on multivariate time series data for merchant transactions.
We design the model with the consideration of learning not only the numerical values but also the shape, the magnitude, and the offsets of the time series data.
Our proposed model is flexible as it now predicts multiple possible futures of the merchant transactions.
We conduct experimental evaluations on real-world merchant transaction data to demonstrate its effectiveness.
\bibliographystyle{splncs04}

\begin{thebibliography}{10}
\providecommand{\url}[1]{\texttt{#1}}
\providecommand{\urlprefix}{URL }
\providecommand{\doi}[1]{https://doi.org/#1}

\bibitem{box1970time}
Box, G., et~al.: Time series analysis forecasting and control holden-day: San
  francisco. BoxTime Series Analysis: Forecasting and Control Holden Day1970
  (1970)

\bibitem{breiman2001random}
Breiman, L.: Random forests. Machine learning  \textbf{45}(1),  5--32 (2001)

\bibitem{cerqueira2017arbitrated}
Cerqueira, V., et~al.: Arbitrated ensemble for time series forecasting. In:
  Joint European conference on machine learning and knowledge discovery in
  databases. pp. 478--494. Springer (2017)

\bibitem{de2018multivariate}
De~Stefani, J., , et~al.: A multivariate and multi-step ahead machine learning
  approach to traditional and cryptocurrencies volatility forecasting. In: ECML
  PKDD 2018 Workshops. pp. 7--22. Springer (2018)

\bibitem{faloutsos2019forecasting}
Faloutsos, C., et~al.: Forecasting big time series: Theory and practice. In:
  ACM SIGKDD (2019)

\bibitem{fan2019multi}
Fan, C., et~al.: Multi-horizon time series forecasting with temporal attention
  learning. In: ACM SIGKDD (2019)

\bibitem{guzman2012multiple}
Guzman-Rivera, A., et~al.: Multiple choice learning: Learning to produce
  multiple structured outputs. In: NeurIPS (2012)

\bibitem{ho1995random}
Ho, T.K.: Random decision forests. In: Proceedings of 3rd international
  conference on document analysis and recognition. vol.~1, pp. 278--282. IEEE
  (1995)

\bibitem{kingma2014adam}
Kingma, D.P., Ba, J.: Adam: A method for stochastic optimization. arXiv
  preprint arXiv:1412.6980  (2014)

\bibitem{lee2016stochastic}
Lee, S., et~al.: Stochastic multiple choice learning for training diverse deep
  ensembles. In: NeurIPS (2016)

\bibitem{radford2015unsupervised}
Radford, A., et~al.: Unsupervised representation learning with deep
  convolutional generative adversarial networks. arXiv preprint
  arXiv:1511.06434  (2015)

\bibitem{rakthanmanon2012searching}
Rakthanmanon, T., et~al.: Searching and mining trillions of time series
  subsequences under dynamic time warping. In: ACM SIGKDD (2012)

\bibitem{babu2018divide}
Sam, B., et~al.: Divide and grow: Capturing huge diversity in crowd images with
  incrementally growing cnn. In: IEEE CVPR (2018)

\bibitem{sen2019think}
Sen, R., et~al.: Think globally, act locally: A deep neural network approach to
  high-dimensional time series forecasting. In: NeurIPS (2019)

\bibitem{shih2019temporal}
Shih, S.Y., et~al.: Temporal pattern attention for multivariate time series
  forecasting. Machine Learning  \textbf{108}(8-9),  1421--1441 (2019)

\bibitem{taieb2015bias}
Taieb, S.B., et~al.: A bias and variance analysis for multistep-ahead time
  series forecasting. IEEE transactions on neural networks and learning systems
   \textbf{27}(1),  62--76 (2015)

\bibitem{tang2019multiple}
Tang, C., et~al.: Multiple futures prediction. In: NeurIPS (2019)

\bibitem{wen2017multi}
Wen, R., et~al.: A multi-horizon quantile recurrent forecaster. arXiv preprint
  arXiv:1711.11053  (2017)

\end{thebibliography}

\end{document}